\documentclass[11pt]{article}
\usepackage[utf8]{inputenc}
\usepackage[T1]{fontenc}
\usepackage{lmodern}
\usepackage[margin=1in]{geometry}
\usepackage{amsmath,amssymb}
\usepackage{graphicx}
\usepackage{booktabs}
\usepackage{caption}
\usepackage[hidelinks]{hyperref}
\usepackage{microtype}
\usepackage[numbers,sort&compress]{natbib}
\usepackage{placeins}
\usepackage{framed}

\title{\bfseries Explanation Multiplicity: Circuit-Level Interpretability\\
Evidence Does Not Survive Defensible Analytic Variation}
\author{Ajay Pravin Mahale}
\date{12 August 2026}

\begin{document}
\maketitle

\begin{abstract}
\noindent
The EU AI Act requires providers of high-risk systems to file technical
documentation describing how the system reaches its decisions. Mechanistic
interpretability is the obvious source of such evidence, and circuit discovery is
its most developed instrument. We ask whether that evidence survives the condition
under which it would be relied upon: two competent analysts, the same system, the
same tool, different defensible settings.

We pre-registered a crossed grid of seven analytic axes, every level taken from a
published implementation, and mapped each discovered circuit through a
deterministic claim map to a structured Annex~IV statement. Across 15{,}840
pre-registered specifications on GPT-2 small and the indirect object
identification task, of which 7{,}561 produced a claim, the derived statement
flips across 73.2\% of specification pairs (95\% CI 0.725 to 0.738) and the modal
claim commands 41.1\% of the space. The evidence fails a filability criterion at
every tolerance a conformity assessment body would plausibly accept.

Standardising the single most influential choice, the evaluation metric, leaves
the flip rate at 59.4\%. Removing circuit size from the claim entirely and holding
it fixed leaves 27.1\% (95\% CI 0.255 to 0.286), still above the pre-registered
threshold. The circuits underlying these claims are structurally near-disjoint,
median pairwise Jaccard overlap 4\%, and functionally uncorrelated at Cohen's
kappa 0.015, so the instability is not one mechanism described in different words.

We give the filability criterion as a standalone protocol, and we report that one
of the seven documented discovery objectives does not execute at all on the
library's own canonical task. The study covers one model and one task, and whether
the conclusion holds at scale is untested.
\end{abstract}

\section{Introduction}

An audit means two auditors reach the same conclusion. That is not a definition
borrowed from machine learning; it is what makes an audit worth commissioning.

Regulation (EU) 2024/1689 requires providers of high-risk AI systems to file
technical documentation. Annex~IV point 2(b) asks for ``the general logic of the
AI system and of the algorithms''. Point 2(e) asks for ``an assessment of the
technical measures needed to facilitate the interpretation of the outputs''.
Article 86(1) gives a person subject to certain automated decisions the right to
``clear and meaningful explanations of the role of the AI system in the
decision-making procedure and the main elements of the decision taken''. These are
quoted from the consolidated text as amended by Regulation (EU) 2026/1744,
CELEX 02024R1689-20260727, read 11 August 2026.

Mechanistic interpretability offers to supply the technical content. Circuit
discovery produces a subgraph said to explain a behaviour, and it is the part of
the field with mature tooling. If a provider files a circuit-level account of why
its system behaved as it did, the filing is worth something only if a second
analyst, given the same system and the same tool, would file something compatible.

We measure how far a regulatory claim moves across the space of defensible
analytic specifications, and attach a decision rule a standards body could use. We
do not propose a better circuit discovery method, and we do not argue that
interpretability is worthless. We argue that the evidence, as currently produced,
does not meet the standard that filing it presumes.

\paragraph{Contributions.}
(i) A pre-registered multiverse over seven analytic axes of circuit discovery,
with every level drawn from a published implementation, and a deterministic map
from circuits to Annex~IV statements.
(ii) A filability criterion, $\pi^{*} \geq 1 - \alpha$, that a conformity
procedure can apply, together with the measured value on a canonical task.
(iii) Evidence that the instability survives the most aggressive controls
available: standardising any single axis, removing circuit size from the claim,
and holding size fixed.
(iv) A methodological result for multiverse designs generally: when
specifications select objects of different sizes, pooled statistics can invert,
and we show three of ours that do.

\section{Related work}

\paragraph{Non-identifiability.} \citet{meloux2025} ask whether, for a given
behaviour and under mechanistic interpretability's own criteria, a unique
explanation exists, and answer that it need not. Our premise is theirs. Our
contribution is downstream: we measure what non-identifiability costs when an
explanation is filed rather than published.

\paragraph{Faithfulness metric sensitivity.} \citet{miller2024} survey design
choices in ablation-based faithfulness measurement and find existing methods
highly sensitive to seemingly insignificant changes, concluding that faithfulness
scores reflect the researcher's methodological choices as well as the circuit.
They establish that one axis matters. We cross seven, and propagate the result to
a regulatory statement. Their released library is the instrument we use, which
means our study is a stress test of the tooling by the tooling's own definitions.

\paragraph{Structure versus function.} \citet{bayatmakou2026} vary input
statistics while holding the task fixed and find that structurally distinct
circuits implement the same computation, a pattern they term phantom
specialisation. This is the strongest objection to our design: if discovery merely
samples from an equivalence class of behaviourally identical subgraphs, a flip
rate over claims measures which member was sampled. We address it with a
functional measurement rather than an argument in Section~\ref{sec:functional},
and the answer is that our circuits are not functionally interchangeable.

\paragraph{Variance as estimation.} \citet{meloux2025b} treat mechanistic
interpretability as statistical estimation and analyse the variance of the
estimator. Seed variance is one axis of our space; their framing is the natural
statistical companion to what we do at the level of the whole specification.

\paragraph{Auditability.} \citet{lan2026} observe that two papers reached
conflicting conclusions about the same behaviour, with a third finding both partly
correct and incomparable, and call for auditing standards so that findings can be
certified in safety-critical settings. We answer that call with a measurement and
a criterion rather than a proposal. \citet{sharkey2025} list open problems in the
field, including the socio-technical ones this paper touches, and
\citet{mueller2025} argue that comparison across methods requires standardised
evaluation, which is adjacent to our argument that certification requires
standardised specification.

\paragraph{Multiverse method.} The design follows \citet{steegen2016}, who propose
reporting results across the space of defensible data-processing choices, and
\citet{simonsohn2020}, whose specification curve is the visualisation we adapt.
\citet{simmons2011} named the underlying problem: undisclosed flexibility in
analysis lets an analyst present almost anything as significant. We are applying a
psychology methods literature to a machine learning evidentiary question, and the
transfer is not decorative. The claim that an explanation is stable is exactly the
kind of claim that flexibility can manufacture.

\paragraph{Explanation metrics under optimisation pressure.}
\citet{hsia2023} show that sufficiency and comprehensiveness can be inflated
without changing predictions or explanations. Two of our four evaluation metrics
are those two, which is one reason we report the discard rate per metric rather
than pooling silently.

\section{What this paper is not claiming}

Stated early because each is a reading the results do not support, and two are
readings we held ourselves before the decompositions were run.

We are not claiming that discovered circuits are no better than random. At fixed
circuit size they are more stable than a size-matched random null, 0.2746 against
0.4230. They carry information.

Nor that filings differ where mechanisms do not. Mechanisms differ a great deal:
functional instability is 0.3972 pooled and 0.4777 at the smallest circuits.

Nor that discovery samples one mechanism and dresses it differently. Cohen's kappa
of 0.0146 says the circuits are functionally uncorrelated.

Nor that random seeds are the problem. The seed here selects which prompts are
sampled, so it is evaluation-set variability rather than nondeterminism.

And nothing at all about scale. One model, one task.

\section{Formalism}

A \emph{specification} is a tuple of analytic choices
$s = (o, a, d, m, \tau, P, r, g)$, with $o$ the discovery objective, $a$ the
ablation operator, $d$ the corruption distribution, $m$ the evaluation metric,
$\tau$ the size threshold, $P$ the prompt variant, $r$ the seed and $g$ the
granularity. Every level of every axis is taken from a published implementation.
None was invented for this study.

Circuit discovery returns $C(s) \subseteq E$, a subset of the model's edge set.
For GPT-2 small under the factorised graph with separate query, key and value
inputs, $|E| = 32{,}491$. We confirm this from the instrument and independently
from a closed-form count over the computation order: block $b$ attention receives
from $1 + 13b$ sources at three input slots for each of 12 heads, MLP $b$ receives
from those sources plus its own block's heads, and the output receives from all
157 non-terminal sources. Circuits touch components drawn from 144 attention heads
and 12 MLPs, so 156 in total.

\paragraph{Circuit instability.} For two specifications,
\begin{equation}
D(s_i, s_j) = 1 - \frac{|C_i \cap C_j|}{|C_i \cup C_j|},
\end{equation}
and $\bar{J}$ is the mean Jaccard similarity over unordered pairs.

\paragraph{Claim instability.} Let $\varphi$ map a circuit to a structured
Annex~IV statement. The flip rate is the probability that two distinct
specifications yield different statements,
\begin{equation}
F = 1 - \frac{\sum_c n_c (n_c - 1)}{N(N-1)},
\label{eq:flip}
\end{equation}
where $n_c$ counts specifications yielding claim class $c$ and $N$ is the number
of specifications. Equation~\ref{eq:flip} is the unbiased Gini--Simpson form. It is
the probability that two draws without replacement differ, and it is unit-tested
against the brute-force pairwise definition over randomised cases. The modal share
is $\pi^{*} = \max_c n_c / N$.

We also use the within-group form. For a set of conditioning axes $G$,
\begin{equation}
F_{\text{within}}(G) = 1 - \frac{\sum_g \sum_c n_{g,c}(n_{g,c}-1)}{\sum_g n_g(n_g-1)},
\label{eq:within}
\end{equation}
which weights every pair equally rather than every group equally, and reduces
exactly to Equation~\ref{eq:flip} when $G$ is empty.

\paragraph{Filability.} Evidence is \emph{filable at tolerance} $\alpha$ if and
only if $\pi^{*} \geq 1 - \alpha$.

Note the ceiling: $F \leq 1 - 1/k$ for $k$ claim classes, so a reported flip rate
must be read against the number of classes the map can produce. With the nine
classes observed here the ceiling is 0.8889, and the observed 0.7316 is 82.3\% of
it. $F$ is not near one by construction.

\section{The claim map}

$\varphi$ is the most attackable choice in the paper, so it is fixed in the
pre-registration, implemented as deterministic code, and published. It is not
generated by a language model; doing so would measure model variance on top of
circuit variance with no way to separate them.

There are two addressees, following the regulation rather than convenience.
$\varphi_{\text{overseer}}$ is built on Annex~IV 2(e) and 3 with Article 14(4)(c),
and keyed on the dominant layer band, then the size class, then the ranked input
segment. $\varphi_{\text{affected}}$ is built on Article 86(1) and leads with the
input segment, because the right is owed to the person the decision is about.
Three granularities are nested by construction, so a claim at a finer granularity
refines rather than contradicts the coarser one.

A reviewer will say the size bins were chosen to produce the reported flip rate.
They were returned by a committed calibration rule from a measured node-count
curve, not chosen. Table~\ref{tab:bins} reports the sensitivity.

\begin{table}[htbp]
\centering
\caption{Flip rate under alternative size binnings. The committed bins are not a
maximum: four alternatives give a higher $F$, including halving every threshold.
Across binnings differing by a factor of four the range is 0.70 to 0.77, and
removing the size term from the claim entirely still leaves 0.5733.}
\label{tab:bins}
\begin{tabular}{lc}
\toprule
Binning & $F$ \\
\midrule
committed bins & 0.7316 \\
all thresholds $\times$ 0.5 & 0.7448 \\
all thresholds $\times$ 0.8 & 0.7435 \\
all thresholds $\times$ 0.9 & 0.7394 \\
all thresholds $\times$ 1.1 & 0.7289 \\
all thresholds $\times$ 1.25 & 0.7314 \\
all thresholds $\times$ 1.5 & 0.7308 \\
all thresholds $\times$ 2 & 0.7232 \\
equal thirds & 0.7018 \\
deciles & 0.7719 \\
no size term (COARSE) & 0.5733 \\
\bottomrule
\end{tabular}
\end{table}

\section{Method}

\paragraph{Model and task.} GPT-2 small on indirect object identification
\citep{wang2022}. The choice is not an apology. The non-identifiability and
faithfulness literature this paper extends works at this scale, comparability is
the point, and the edge count is what makes a size-matched random baseline
meaningful.

\paragraph{Grid.} Seven discovery objectives, seven ablation operators, four
corruption distributions nested within the five operators that read them, four
metrics, three thresholds, two prompt variants, five seeds: 1{,}540 discovery
cells and 18{,}480 specifications as pre-registered. Corruption is nested rather
than crossed because two ablation operators never read the corrupt distribution,
so crossing them would emit duplicate specifications. Discovery uses the edge
attribution patching family \citep{syed2023} and integrated edge gradients, as
implemented in the library released with \citet{miller2024}; the automated circuit
discovery lineage is \citet{conmy2023}. The four corruption levels are taken from
\citet{wang2022}.

\paragraph{Pre-registration.} The analysis plan was committed and tagged before
the confirmatory sweep began, with both abstracts drafted in advance so that
neither direction could be written up as a surprise. Deviations are recorded in an
append-only file with their dates and whether each was decided before or after the
affected result was seen.

\paragraph{Statistics.} No $p$-value is computed across specifications anywhere.
Specifications are a designed grid, not an independent sample
\citep{steegen2016,simonsohn2020}. Uncertainty comes from a nonparametric
bootstrap that resamples specifications, never pairs, at $B = 10{,}000$, because
$F$ and $\bar{J}$ are U-statistics over pairs and pairs sharing a specification
are dependent. Intervals quantify uncertainty given the grid and license nothing
beyond it.

\paragraph{Two exclusions, criteria fixed before any circuit was seen.} One
discovery objective does not execute. All 220 of its cells raise a runtime error
from a mean squared error computed against an integral target, inside the
library's own code. We verified by digest over the sorted failing cell
identifiers that the failure set is exactly that arm and contains nothing else. We
did not repair it: the library is the instrument under test and is reproduced
verbatim. The realised grid is six objectives, 1{,}320 cells and 15{,}840
specifications.

The metric-relative threshold rule admits no circuit for 8{,}279 of 15{,}840
specifications, a discard rate of 52.3\%, distributed unevenly across metrics
(Table~\ref{tab:discard}). This was disclosed before the lock. The surviving pool
is therefore 52\% comprehensiveness against 25\% by design, and we report the
composition rather than smoothing it.

\begin{table}[htbp]
\centering
\caption{Discard rate by evaluation metric under the pre-registered
metric-relative threshold rule. The unevenness was disclosed before the
pre-registration was locked, from a six-cell validation slice.}
\label{tab:discard}
\begin{tabular}{lccc}
\toprule
Metric & Kept & Discarded & Rate \\
\midrule
comprehensiveness & 3,947 & 13 & 0.003 \\
KL divergence & 1,467 & 2,493 & 0.630 \\
logit difference & 1,270 & 2,690 & 0.679 \\
sufficiency & 877 & 3,083 & 0.779 \\
\bottomrule
\end{tabular}
\end{table}

\section{Results}

Table~\ref{tab:hyp} states each pre-registered outcome against its result. The
decision rules and thresholds were fixed and tagged before the confirmatory sweep
ran, and none was revisited.

\begin{table}[htbp]
\centering
\caption{Pre-registered outcomes and their results. Thresholds were fixed on
6 August 2026 and the plan was tagged before the sweep began. Two results carry
qualifications that only the within-size decomposition reveals, and those are
reported in Sections~\ref{sec:null} and \ref{sec:functional} rather than deferred
to an appendix.}
\label{tab:hyp}
\begin{tabular}{llll}
\toprule
& Decision rule & Result & Outcome \\
\midrule
P0 & $\bar{J}$ reported, not tested & 0.1396 [0.1377, 0.1418] & premise holds \\
H2 & $F > 0.20$, CI lower bound above & 0.7316 [0.7247, 0.7380] & \textbf{confirmed} \\
H3 & not separated from random null & separated, intervals disjoint & \textbf{rejected} \\
H4 & $F - (1 - \text{agreement}) > 0.10$ & 0.3344 [0.3297, 0.3394] & \textbf{supported} \\
\bottomrule
\end{tabular}
\end{table}

\subsection{Circuits are almost disjoint}

$\bar{J} = 0.1396$ (95\% CI 0.1377 to 0.1418), computed exactly over all
28{,}580{,}580 specification pairs rather than sampled. The 7{,}561
specifications hold 3{,}218 distinct circuits, which makes the exact computation
tractable.

The distribution says more than the mean. Quantiles of pairwise $D$: the 1\%
quantile is 0.196, the 25\% quantile 0.788, the median 0.960, and the 75\%
quantile 0.998. The median pair of circuits shares 4\% of its edges, and only 1\%
of pairs exceed a Jaccard of 0.8. At the median circuit size of 200 edges the
analytic random line $k/(2|E|-k)$ gives 0.0031, so observed overlap is about 45
times chance. Beating chance at this scale is not evidence of much: a 500-edge
circuit has an expected random overlap of 0.008.

\subsection{The claim flips on three quarters of pairs}

$F = 0.7316$ (95\% CI 0.7247 to 0.7380) against a pre-registered threshold of 0.20
with the interval lower bound required to clear it. It clears by more than three
times. $\pi^{*} = 0.4109$, so the evidence is not filable at any pre-registered
tolerance: 0.411 against requirements of 0.95, 0.90 and 0.80.

\begin{table}[htbp]
\centering
\caption{The nine claim classes produced by $\varphi_{\text{overseer}}$ at medium
granularity, ordered by frequency. Aggregating over size class, the layer
attribution splits 47.9\% early, 43.6\% late and 8.5\% middle; the first of these
is the COARSE modal share of 0.4788, computed independently.}
\label{tab:classes}
\begin{tabular}{clccc}
\toprule
Rank & Layer band & Size class & Count & Share \\
\midrule
1 & early layers & distributed & 3,107 & 0.4109 \\
2 & late layers & sparse & 2,153 & 0.2848 \\
3 & late layers & moderate & 657 & 0.0869 \\
4 & late layers & distributed & 489 & 0.0647 \\
5 & early layers & sparse & 477 & 0.0631 \\
6 & middle layers & moderate & 283 & 0.0374 \\
7 & middle layers & sparse & 269 & 0.0356 \\
8 & middle layers & distributed & 90 & 0.0119 \\
9 & early layers & moderate & 36 & 0.0048 \\
\bottomrule
\end{tabular}
\end{table}

The two most common claims, at 41.1\% and 28.5\% of the space, attribute the same
model's behaviour on the same task to early layers and to late layers
(Table~\ref{tab:classes}). They are not different emphases. They contradict.

\begin{figure}[htbp]
\centering
\includegraphics[width=0.92\textwidth]{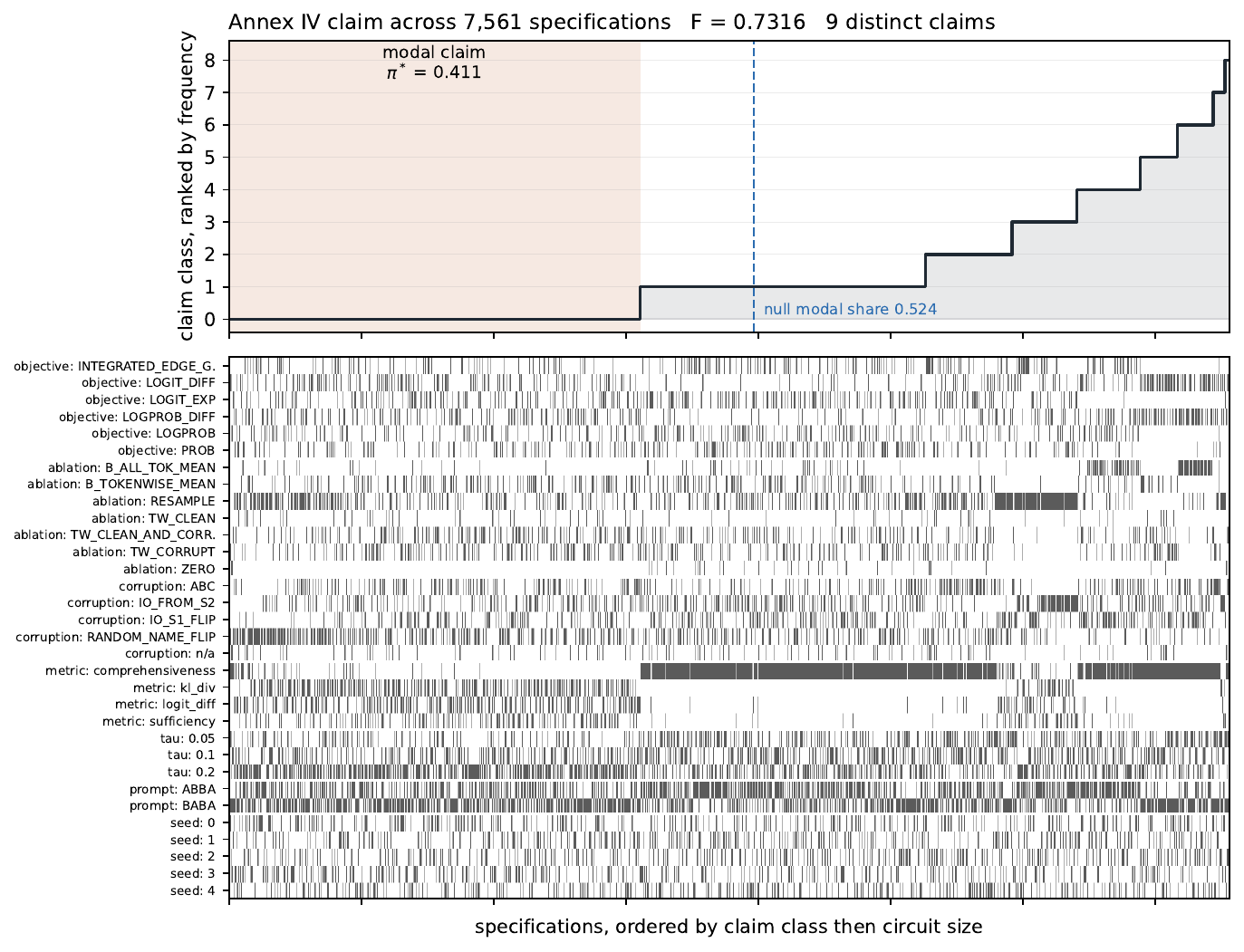}
\caption{Specification curve for $\varphi_{\text{overseer}}$ at medium
granularity. Upper panel: claim class against specification rank, classes ordered
by frequency, so each step's width is that class's share of the space. The dashed
line marks the size-matched random null's modal share; the pooled comparison is
confounded by circuit size, discussed in Section~\ref{sec:null}. Lower panel:
which analytic level is active for each specification. The solid band on the
comprehensiveness row is the composition effect of the discard rule in
Table~\ref{tab:discard}; the seed rows are unstructured, which is the same result
as seed removing 0.0009 from $F$ in Table~\ref{tab:fixed}.}
\label{fig:speccurve}
\end{figure}

All six addressee-granularity combinations reject filability
(Table~\ref{tab:gran}). The coarsest statement available to an affected person is
a two-class claim and it still flips on 39.5\% of pairs, so the result is not an
artefact of fine-grained claim language. Figure~\ref{fig:speccurve} shows the
specification curve; the width of the modal step is $\pi^{*}$.

\begin{table}[htbp]
\centering
\caption{Flip rate and modal share for both addressees at all three
granularities. Filability is evaluated at the loosest pre-registered tolerance,
$\alpha = 0.20$, which requires $\pi^{*} \geq 0.80$. No cell passes.}
\label{tab:gran}
\begin{tabular}{llcccc}
\toprule
Addressee & Granularity & Classes & $F$ & $\pi^{*}$ & Filable \\
\midrule
affected & COARSE & 2 & 0.3950 & 0.7291 & no \\
affected & MEDIUM & 4 & 0.6597 & 0.4875 & no \\
affected & FINE & 8 & 0.6744 & 0.4786 & no \\
overseer & COARSE & 3 & 0.5733 & 0.4788 & no \\
overseer & MEDIUM & 9 & 0.7316 & 0.4109 & no \\
overseer & FINE & 12 & 0.7681 & 0.4109 & no \\
\bottomrule
\end{tabular}
\end{table}

\subsection{No analytic axis rescues the filing}

\begin{table}[htbp]
\centering
\caption{Residual flip rate once each axis is standardised, computed with
Equation~\ref{eq:within}. All intervals are valid: group sizes 1{,}512 to 2{,}520,
bootstrap bias below 0.0006 on every axis, and every observed value lies inside
its own interval.}
\label{tab:fixed}
\begin{tabular}{lccc}
\toprule
Axis standardised & Residual $F$ & 95\% CI & Removes \\
\midrule
Evaluation metric & 0.5939 & 0.5803 to 0.6063 & 0.1377 \\
Ablation operator & 0.7014 & 0.6939 to 0.7075 & 0.0302 \\
Discovery objective & 0.7018 & 0.6938 to 0.7084 & 0.0298 \\
Threshold $\tau$ & 0.7070 & 0.6981 to 0.7149 & 0.0246 \\
Corruption & 0.7092 & 0.7008 to 0.7166 & 0.0224 \\
Prompt variant & 0.7232 & 0.7157 to 0.7300 & 0.0084 \\
Seed & 0.7307 & 0.7234 to 0.7368 & 0.0009 \\
\bottomrule
\end{tabular}
\end{table}

The metric is the largest single lever and it is not enough. Standardising it
completely, which is more than a conformity procedure could demand, leaves
$F = 0.5939$ with the interval lower bound at 0.5803, still far above threshold.

Structure and claim are governed by different axes. A REML variance decomposition
on $\log_{10}$ of the selected circuit size attributes 78.8\% of structural
variance to the ablation-corruption interaction, 11.5\% to residual, 4.4\% to
ablation, 4.1\% to the threshold, and 0.7\% to the metric, which is the largest
driver of claim variance. A standards body reading the structural decomposition
would standardise the ablation operator and fix almost none of the filing
instability.

\subsection{The result under maximum pressure}

Strip every trace of circuit size from the claim, so the statement is the dominant
layer band alone, and hold circuit size fixed with Equation~\ref{eq:within}:
$F = 0.2706$ (95\% CI 0.2550 to 0.2859). The interval lower bound is above the
pre-registered threshold. Where in the model the behaviour is attributed still
flips on 27\% of pairs when nothing about size can contribute.

\subsection{The random baseline, and why its pooled direction misleads}
\label{sec:null}

Pooled, the discovered multiverse is \emph{less} stable than a size-matched random
null drawn from the full 32{,}491-edge population, $R = 1{,}000$ null multiverses:
0.7316 against 0.6774, intervals disjoint.

H3 is therefore rejected: the distributions are separated. We do not report the
direction of that separation as a finding, because it is an artefact of the size
distribution, and Table~\ref{tab:bysize} shows why. The null is degenerate above 1{,}000 edges,
where a uniform draw touches all 156 components and can produce only one claim.
With size held fixed the ordering reverses: discovered 0.2746 against null 0.4230
at medium granularity, and 0.2706 against 0.3822 at coarse. Discovered circuits
are more stable than random ones once the comparison is fair. They are still not
stable enough to file.

\begin{table}[htbp]
\centering
\caption{Flip rate within each circuit size, medium granularity. The null
collapses to a single claim class at and above 1{,}000 edges, which is what
depresses its pooled flip rate. The discovered multiverse also collapses at
5{,}000 and 10{,}000 edges, so 2{,}692 of 7{,}561 specifications contribute no
flips at all and the instability lives entirely below 2{,}000 edges.}
\label{tab:bysize}
\begin{tabular}{rrcccc}
\toprule
Size & $n$ & Discovered $F$ & Classes & Null $F$ & Null classes \\
\midrule
10 & 2,492 & 0.4543 & 3 & 0.7331 & 6 \\
20 & 489 & 0.3380 & 4 & 0.6622 & 3 \\
50 & 519 & 0.4639 & 3 & 0.7310 & 6 \\
100 & 242 & 0.5792 & 3 & 0.6584 & 3 \\
200 & 199 & 0.5122 & 4 & 0.6366 & 3 \\
500 & 236 & 0.4258 & 3 & 0.1421 & 3 \\
1,000 & 380 & 0.5133 & 3 & 0.0000 & 1 \\
2,000 & 312 & 0.5714 & 3 & 0.0000 & 1 \\
5,000 & 625 & 0.0000 & 1 & 0.0000 & 1 \\
10,000 & 2,067 & 0.0000 & 1 & 0.0000 & 1 \\
\bottomrule
\end{tabular}
\end{table}

\subsection{Filings differ more than mechanisms, over part of the range}
\label{sec:functional}

Per-example functional agreement between circuits is 0.6028, so functional
instability is 0.3972, and the gap to the claim flip rate is 0.3344 (95\% CI
0.3297 to 0.3394) against a pre-registered threshold of 0.10. Agreement is
computed from per-example correct or incorrect verdicts using the measures of
\citet{bayatmakou2026}, reproduced verbatim.

Cohen's kappa is 0.0146. The 60\% raw agreement is almost entirely explained by
both circuits being right about the same fraction of examples, not the same
examples. Kappa stays between 0.016 and 0.047 at every circuit size while raw
agreement climbs from 0.52 to 0.92, tracking accuracy from 0.51 to 0.95.

This is the answer to the phantom specialisation objection. If discovery sampled
from an equivalence class of behaviourally identical subgraphs, the circuits would
be functionally interchangeable. At kappa 0.015 they are not.

H4 is supported on the pre-registered quantity. The gap does not survive holding
size fixed: it fails the 0.10 threshold at 6 of 10 sizes and is negative at 5. At both ends of the range the mechanisms differ
more than the filings do, and the pooled figure is carried by the 500 to 2{,}000
edge band alone. We report the pooled result because it was pre-registered, and
the decomposition beside it because the pooled result alone would mislead.

\section{The filability protocol}

\begin{framed}
\noindent\textbf{Filability of an interpretability claim.} For a filed claim $c$
derived from a circuit:
\begin{enumerate}\itemsep2pt
\item Declare the specification space $\mathcal{S}$: the analytic axes and the
levels of each that a competent analyst could defend, each with a citation.
\item Declare the claim map $\varphi$ as deterministic, published code. Not a
language model.
\item Compute $\pi^{*} = \max_c n_c / N$ over $\mathcal{S}$, and $F$ by
Equation~\ref{eq:flip}.
\item The claim is filable at tolerance $\alpha$ if $\pi^{*} \geq 1 - \alpha$.
\item Report $\pi^{*}$, $F$, the discard rate per axis level, and the
within-group decomposition of any pooled statistic.
\end{enumerate}
\end{framed}

\noindent
A filing that cannot state its own $\pi^{*}$ has not been audited. It has been
asserted. The protocol does not require the regulator to adjudicate which analytic
choice is correct, which is the property that makes it implementable: it requires
only that the filer declare the space and report how much of it agrees.

\section{Limitations}

\paragraph{One model, one task.} Larger models have more components, so both the
layer band and the size class have more room, and the direction of the effect on
$F$ is not obvious. This is the limitation most likely to change the number.

A second-model replication is pre-registered but not yet run: a 924-cell grid on
Pythia-160m, a model of a different architecture family that performs the same
task, identical on every axis except that the seed axis carries three levels
rather than five. The analysis plan, the reduced grid, and the criteria for what
would count as replication rather than refutation are committed and timestamped
before any cell of it ran. It will be reported in a revision in whichever
direction it falls. A second task remains designed and not run.

\paragraph{Circuit size is an outcome, not an axis.} It is chosen by the threshold
rule, and it has distorted three separate pooled quantities: $F$ itself, the
direction of the null comparison, and the sign of the functional gap. Fixing it
drops $F$ from 0.7316 to 0.2746, a larger reduction than standardising any
pre-registered axis. We take this to be a general hazard for multiverse designs
whose specifications select objects of different sizes, and we report every pooled
statistic with its within-size decomposition.

\paragraph{The multiverse is as wide as the library, not the literature.} Optimal
ablation \citep{li2024} is a defensible operator the instrument does not
implement, so it never entered the space. The measured $F$ is therefore a lower
bound with respect to the field's full set of defensible choices.

\paragraph{The discard rate is high and uneven.} 52.3\%, and the surviving pool is
metric imbalanced. The threshold rule was fixed in advance and no failing item was
repaired, but the composition is a property of the reported set and we state it
rather than adjust for it.

\paragraph{The bootstrap is invalid for one reported family.} Where the
conditioning groups hold only two to five specifications, retained self-pairs
dominate and the intervals are unusable; those quantities are reported as point
estimates with the diagnostic attached. We did not construct a bias-corrected
estimator after seeing the naive intervals disagree, because choosing an estimator
in response to a result is the behaviour this paper documents.

\section{Discussion}

\paragraph{What standardising one axis buys.} Less than it appears. The metric is
the biggest lever and removes 0.1377 of a 0.7316 flip rate. There is no single
knob, and a procedure that standardises one axis and declares the problem solved
would be mistaken by a wide margin.

\paragraph{On the instrument.} One of seven documented, publicly exported
discovery objectives does not run on the library's own canonical task. We report
this as a discard with its rate, and as an observation about the maturity of the
tooling that regulatory evidence is expected to rest on. It is not the paper's
headline, and it is not a criticism of a library whose authors are among those
calling for the field to address exactly this \citep{sharkey2025}.

\paragraph{What a regulator can take from this.} Not that interpretability
evidence is worthless, but that its evidentiary weight is a measurable property
rather than an assumed one, and that the measurement is cheap relative to the
sweep that produced the circuit in the first place.

\section{Reproducibility}

The pre-registration is a timestamped commit tagged \texttt{prereg-p1-confirmatory}
in the public repository, made before the confirmatory sweep began. Every reported
number traces to a config, a seed and an environment fingerprint.
The confirmatory sweep ran under a single environment across all 1{,}320 cells,
verified from the results archive rather than asserted, and the code commit was
recovered by content-addressed comparison of the uploaded archive against the git
tree, 22 of 22 files identical. The analysis layer runs without a GPU.

A verification script recomputes every headline quantity from raw results and
asserts against what is reported: 36 checks, 0 failures, with $F$ derived by three
routes including a brute-force pairwise computation. A second script checks that
every numeric token in this manuscript matches a value derived from the stored
results. Every table in this paper is emitted directly from those results rather
than transcribed. The test suite is 343 tests.

\FloatBarrier

\bibliographystyle{plainnat}

\end{document}